\documentclass[10pt,twocolumn,a4paper]{esaAI}
\newcommand{\mb}[1]{\mathbf{#1}}
\usepackage{siunitx}
\usepackage{amsmath}
\usepackage{amssymb}
\usepackage{wasysym}

\usepackage{tikz}
\usetikzlibrary{calc,patterns,angles,quotes}

\DeclareMathOperator{\diag}{diag}

\title{Guidance and Control Networks with Periodic Activation Functions}

\def\authorEmail{sebastien.origer@esa.int}

\author[1]{Sebastien Origer\thanks{Corresponding author. E-Mail: \authorEmail}}
\author[1]{Dario Izzo}

\affil[1]{Advanced Concepts Team, European Space Research and Technology Centre (ESTEC), Noordwijk, The Netherlands.}

\begin{document}

\makeCustomtitle

\begin{abstract}
Inspired by the versatility of sinusoidal representation networks (SIRENs), we present a modified Guidance \& Control Networks (G\&CNETs) variant using periodic activation functions in the hidden layers. We demonstrate that the resulting G\&CNETs train faster and achieve a lower overall training error on three different control scenarios on which G\&CNETs have been tested previously. A preliminary analysis is presented in an attempt to explain the superior performance of the SIREN architecture for the particular types of tasks that G\&CNETs excel on.
\end{abstract}

\section{Introduction}
The advantages of using periodic functions as network nonlinearities are well-established. The strategic use of trigonometric functions has long been essential for approximating complex phenomena and representing them succinctly, particularly in signal decomposition \cite{serov2017fourier}. More recently, this approach has also gained prominence in machine learning \cite{sitzmann2019siren}.
Guidance \& Control Networks, coined G\&CNETs, have emerged recently as an alternative to traditional Model Predictive Control schemes (MPC) \cite{izzo2023optimality}. 
Typically G\&CNETs take in the current state of the system as input and their output is directly interpretable as the corresponding optimal controls, effectively fusing the guidance and control steps into a single network inference. G\&CNETs have demonstrated the ability to learn the solution to the Bellmann equations, hence capturing the corresponding optimality principle in a single neural network. The vast majority of G\&CNETs in the literature currently apply ReLU, Softplus or hyperbolic tangent as activation functions in their hidden layers \cite{dario_seb_gcnet, robin_thesis, aggressive_online_control, origer2023guidance, origer2024closing, federici2021deep, stability_dario, izzo2021real}. 
We show that when the G\&CNETs are trained via behavioural cloning, that is, over a training dataset of optimal trajectories, using a periodic activation function for the hidden layer results in much more accurate networks. 
Our findings were inspired by the work of Sitzmann et al. \cite{sitzmann2019siren}, who came up with sinusoidal representation networks (SIRENs), which showed very impressive approximation power for image and video reconstruction, as well as complex boundary value problems, surpassing more common activation functions.

We suggest that since the behavioural cloning approach is a simple regression task, the SIREN architecture \cite{sitzmann2019siren} is far better suited for G\&CNETs.
We tested our findings on three test cases where G\&CNETs have been successfully deployed \cite{dario_seb_gcnet,origer2023guidance, origer2024closing}, each possessing distinct dynamics:
\begin{itemize}
    \item \textit{Drone racing}: This problem is characterized by its highly nonlinear dynamics and its large number of state variables (16 G\&CNET inputs). The control policies range from smooth control inputs (energy-optimal flight) to discontinuous 'bang-bang' inputs (time-optimal flight). The typical time-of-flight of an optimal trajectory is on the order of seconds. \cite{origer2023guidance}.
    \item \textit{Asteroid landing}: Here we attempt to perform a mass-optimal landing on the asteroid Psyche. This problem is described by nonlinear equations of motion, seven G\&CNET inputs and a discontinuous 'bang-bang' profile for the throttle control input as well as the thrust direction. A typical optimal trajectory has a time-of-flight on the order of minutes to hours. \cite{origer2024closing}.
    \item \textit{Interplanetary transfer}: The goal here is to learn the optimal thrust direction in order to perform a time-optimal, constant acceleration, low-thrust interplanetary transfer from the asteroid belt to a target circular orbit. This problem is described by nonlinear equations and the G\&CNET takes in six states as input. A typical optimal trajectory has a time-of-flight on the order of years. \cite{dario_seb_gcnet}.
\end{itemize}
In all cases, switching to the SIREN architecture results in a substantial improvement in G\&CNET accuracy.
\section{Methods}\label{sec:method}
We use the SIREN architecture by Sitzmann et al. \cite{sitzmann2019siren}. These are simple feed-forward neural networks with sine as the activation function for the hidden layers:
\begin{equation}\label{eq:SIREN}
\text{SIREN}: \left\{ 
\begin{array}{l}
        \Phi(\mathbf{x}) = \mathbf{W}_n(\phi_{n-1}\circ\phi_{n-2}\circ \dots \circ \phi_0)(\mathbf{x})+\mathbf{b}_n \\
        \mathbf{x}_i \mapsto \phi_i(\mathbf{x}_i)=\sin{(\omega_0\mathbf{W}_i\mathbf{x}_i+\mathbf{b}_i)}
\end{array}
\right.
\end{equation}
where the $i^{th}$ layer of the neural network is denoted by $\phi_i:\mathbb{R}^{M_i}\mapsto\mathbb{R}^{N_i}$. $\mathbf{W}_i\in \mathbb{R}^{N_i \times M_i}$ is the weight matrix, $\mathbf{b}_i\in \mathbb{R}^{N_i}$ is the bias vector and $\mathbf{x}_i\in \mathbb{R}^{M_i}$ is the input vector. Non-linearity is introduced by passing each component of the resulting vector through the sine. We use $\omega_0=30$ as this value works best in our experiments, it is also the suggested value in \cite{sitzmann2019siren}. There are two other crucial steps for successful training of SIREN networks. Ideally, the input should be uniformly distributed in $x\sim \mathcal{U}(-1,1)$, in our experiments we only scaled our training data such that the minimum and maximum value for each input corresponds to $-1$ and $1$, respectively. Additionally, the weights need to be initialized in $w_i\sim \mathcal{U}(-1/n,1/n)$ for the first layer and $w_i\sim \mathcal{U}(-\sqrt{6/n}/\omega_0,\sqrt{6/n}/\omega_0)$ for all subsequent layers where $n$ is the number of inputs. This weight initialization is detailed in \cite{sitzmann2019siren}.
\section{Results}
In this Section we briefly introduce each optimal control problem and the provide a comparison between the traditional G\&CNETs using ReLU and Softplus activation functions for the hidden layer with the new SIREN architecture. For all three problems we use 80\% / 20\% split for training and validation data, the Adam optimizer \cite{kingma2014adam} with a learning rate of $5\cdot10^{-5}$, no weight decay and a scheduler which reduces the learning rate by a factor $0.9$ whenever the loss does not improve over 10 consecutive epochs. 
For the SIREN network we follow the weight initialization scheme described in Sec.\ref{sec:method},  and for all other networks we use the Kaiming Normal initialisation method \cite{he2015delving}. 
\subsection{Drone racing}
We utilize two coordinate frames as illustrated in App.\ref{app:1} along with a quadcopter model that has 16 states $\mb{x} = [\mb{p}, \mb{v}, \boldsymbol{\lambda}, \boldsymbol\Omega, \boldsymbol{\omega}]$ and four control inputs $\mb{u} = [u_1, u_2, u_3, u_4]$ \cite{origer2023guidance}.
The state vector $\mb{x}$ includes the position $\mb{p}=[x,y,z]$ and velocity $\mb{v}=[v_x,v_y,v_z]$, both of which are defined in the world frame.
The Euler angles $\boldsymbol{\lambda}=[\phi, \theta, \psi]$, which define the orientation of the body frame, the angular velocities $\boldsymbol{\Omega}=[p,q,r]$ within the body frame, and the propeller rates $\boldsymbol{\omega}=[ \omega_1, \omega_2, \omega_3, \omega_4]$.
The control inputs $\mb{u} = [u_1, u_2, u_3, u_4]$ are constrained within $u_i \in [0,1]$, where $u_i=0$ and $u_i=1$ correspond to the minimum ($\omega_{min}$) and maximum ($\omega_{max}$) rotational speeds of the propellers, respectively.
The equations of motion are as follows:
\begin{equation}
\label{eq:dyn}
f(\mb{x}, \mb{u}) = \left\{ 
\begin{array}{l}
        \dot{\mb{p}} = \mb{v} \\
        \dot{\mb{v}} = \mb{g} + R(\boldsymbol{\lambda}) \mb{F} \\
        \dot{\boldsymbol\lambda} = Q(\boldsymbol{\lambda}) \mb{\Omega} \\
        I \dot{\boldsymbol{\Omega}} = - \boldsymbol{\Omega} \times I \boldsymbol{\Omega} + \mb{M}+ \mb{M}_{ext}\\
        \dot{\boldsymbol\omega}  = ((\omega_{max}-\omega_{min}) \mb{u} +\omega_{min} - \boldsymbol\omega)/\tau
\end{array}
\right.
\end{equation}
where $I=\diag (I_x, I_y, I_z)$ represents the moment of inertia matrix and $\mb{g} = [0, 0, g]^T$ with $g=9.81$~\si{\meter\per\second\squared} is the gravitational acceleration. The rotation matrix $R(\mb{\boldsymbol\lambda})$, forces $\mb{F}$, moments $\mb{M}$, and model coefficients are detailed in App.\ref{app:1}.
We use the following cost function $J(\mathbf{u}, T) = (1-\epsilon)T + \epsilon \int_{0}^{T} ||\mb{u}(t)||^2 dt$ to minimize two goals: the total flight time $T$ and the total energy consumption $\int_{0}^{T} ||\mb{u}(t)||^2 dt$. Both objectives are balanced by the hybridization parameter $\epsilon$. $\epsilon=1$ corresponds to energy-optimal flight and $\epsilon=0$ time-optimal flight. 
Although not directly aimed at minimizing time, the energy-efficient component contributes to smoother control inputs, allowing for more tolerance to errors compared to the strictly time-optimal 'bang-bang' control profile.
Denoting $X$ as the state space and $U$ as the set of permissible controls, the optimal control problem at hand seeks to determine the optimal control policy $\mb{u} : [0,1] \rightarrow U$. This policy aims to guide the quadcopter from initial conditions $\mb{x}_0$ to a defined set of final conditions $S$, all while minimizing the cost function $J(\mb{u}, T)$.
For brevity, we omit the specifics of solving this optimal control problem using a direct method and refer interested readers to \cite{origer2023guidance}.
In all cases we use 10,000 optimal trajectories sampled at 199 points. We train the networks for 50 epochs with a training batchsize of 1024 optimal state-action pairs $(\mb{x}^*, \mb{u}^*)$. 
The G\&CNETs are composed of an input layer which takes in 16 inputs (the state vector of the drone), three hidden layers with 128 neurons and an output layer with a Sigmoid activation function which bounds the control inputs between 0 and 1. In total, a G\&CNET has 37,636 parameters. 
Finally, we use the mean squared error (MSE) as the loss function $\mathcal \mathcal{L}=|| \text{G\&CNET}(\mb{x}^*) - \mb{u}^* ||^2$.

\begin{figure}[!t]
  \centering
  \includegraphics[width=\columnwidth]{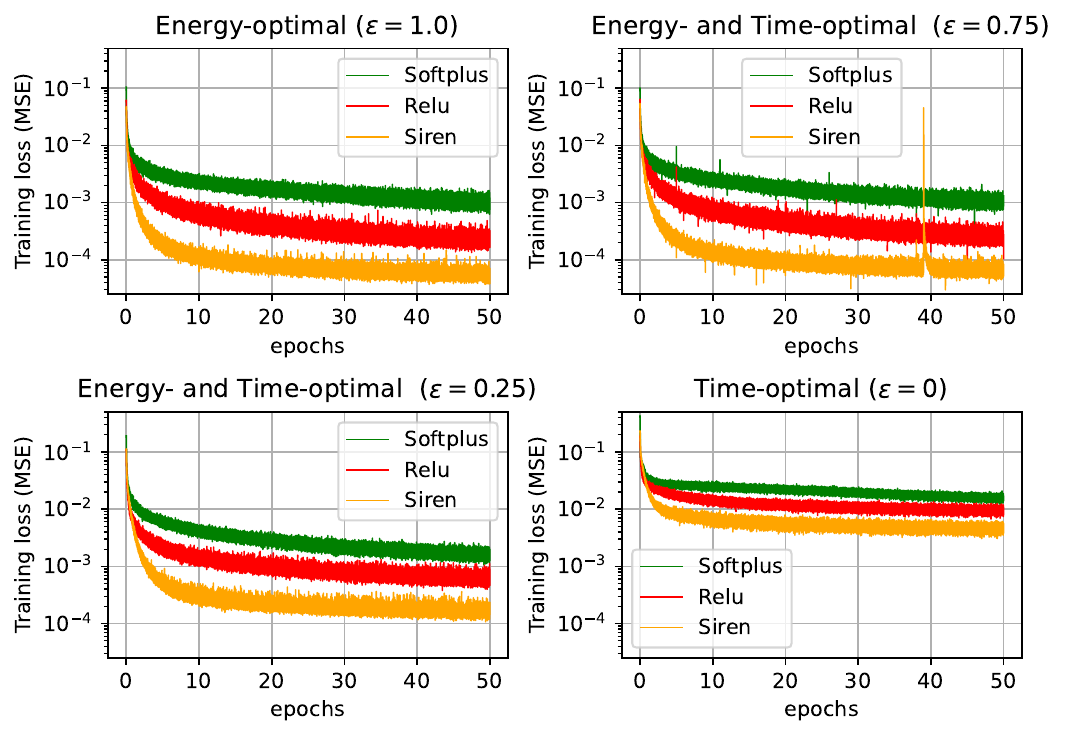}
  \caption{Drone racing: Training loss for different control policies and hidden layer activation functions.}
	\label{fig:quad_losses}
\end{figure}

We report the training loss for four different control policies ranging from fully energy-optimal to fully time-optimal in Fig.\ref{fig:quad_losses}. 
In all cases, the SIREN network converges more quickly and to a lower final training loss compared to the other networks. The smoothness of the optimal control inputs also does not seem to alter this trend, SIRENs perform best both for very smooth ($\epsilon=1.0$) and discontinuous 'bang-bang' ($\epsilon=0.0$) control profiles. Note that in the case of drones, the noise encountered onboard is usually much larger \cite{izzo2023optimality} compared to the space-related problems in Subsec.\ref{subsec:ast} and \ref{subse:transfer}. 
To handle this noise, the G\&CNET requires a high onboard inference frequency, possibly creating a trade-off between low training loss and inference speed.
ReLU was used as activation function in \cite{origer2023guidance} because it offered the quickest inference time. The full version of this paper will include further tests in simulation or onboard a real drone to assess the potential increase in flight performance one might expect from a lower training loss.

\subsection{Asteroid landing}\label{subsec:ast}
Let $\mathcal R = [\hat{\mathbf i}, \hat{\mathbf j}, \hat{\mathbf k}]$ denote a rotating frame with angular velocity $\boldsymbol \omega \hat {\mathbf k}$, ensuring that the asteroid remains stationary within $\mathcal R$ \cite{origer2024closing}.
The equations of motion are:
\begin{equation}
\label{eq:dyn_landing}
\left\{ 
\begin{array}{l}
    \dot{x} = v_x \\
    \dot{y} = v_y \\
    \dot{z} = v_z \\
    \dot{v}_x = -\frac{\mu}{r^3}x + 2 \omega v_y + \omega^2 x +u\frac{c_1}{m} i_x \\
    \dot{v}_y = -\frac{\mu}{r^3}y - 2 \omega v_x + \omega^2 y +u\frac{c_1}{m} i_y \\
    \dot{v}_z = -\frac{\mu}{r^3}z +u\frac{c_1}{m} i_z \\
    \dot{m} = -u\frac{c_1}{I_{sp}g_0}
\end{array}
\right.
\end{equation}
The state vector $\mathbf{x}$ comprises the position $\mathbf{r}=[x,y,z]$, velocity $\mathbf{v}=[v_x,v_y,v_z]$, and mass $m$ of the spacecraft.
Both the position $\mathbf{r}$ and velocity $\mathbf{v}$ are defined within the rotating frame $\mathcal R$.
Note that $r = \sqrt{x^2+y^2+z^2}$ and $\mu$ is the gravitational constant of Psyche.
The system is governed by the thrust direction specified by the unit vector $\mathbf{\hat{t}} = [i_x, i_y, i_z]$ and the throttle $u \in [0,1]$.
We consider a mass-optimal and free-time control problem, wherein we seek to determine the controls $u(t)$ and $\mathbf{\hat{t}}(t)$, with $t\in [t_0,t_f]$. These controls should steer the system, described by Eq.\ref{eq:dyn_landing}, from any initial state $\mathbf{r}_0$, $\mathbf{v}_0$, $m_0$ to the desired target state $\mathbf{r}_t$, $\mathbf{v}_t$. Note that the final mass $m_f$ is left free. We are thus minimizing the following cost function: $J = m_0-m_f$. We solve this problem using an indirect method (Pontryagin's Maximum Principle \cite{pontryagin}), for details on the exact implementation we refer the reader to \cite{origer2024closing}. 
A portion of the training dataset is shown in App.\ref{app:2}.

For all G\&CNETs we use 300,000 optimal trajectories sampled at 100 points. We train the networks for 500 epochs with a training batchsize of 2048 optimal state-action pairs $(\mb{x}^*, \mb{u}^*)$.
The G\&CNETs are composed of an input layer which takes in 7 inputs (the state vector of the spacecraft), three hidden layers with 128 neurons each, and a linear output layer. The throttle input is passed through a Sigmoid activation function to bound it between 0 and 1. The resulting G\&CNET has 34,564 parameters in total.
Finally, we use a loss function of the form $\mathcal L= \text{MSE}(u_{nn},u^*) + 1-\frac{\mathbf{\hat{t}}^*\cdot \mathbf{\hat{t}}_{nn}}{\mathbf{\vert\hat{t}}^*\vert\vert \mathbf{\hat{t}}_{nn}\vert}$, where the network learns to minimize the mean squared error between the estimated throttle $u_{nn}$ and the ground truth $u^*$ and the cosine similarity of the estimated thrust direction $\mathbf{\hat{t}}_{nn}$ and the ground truth $\mathbf{\hat{t}}^*$.
\begin{figure}[!]
  \centering
  \includegraphics[width=\columnwidth]{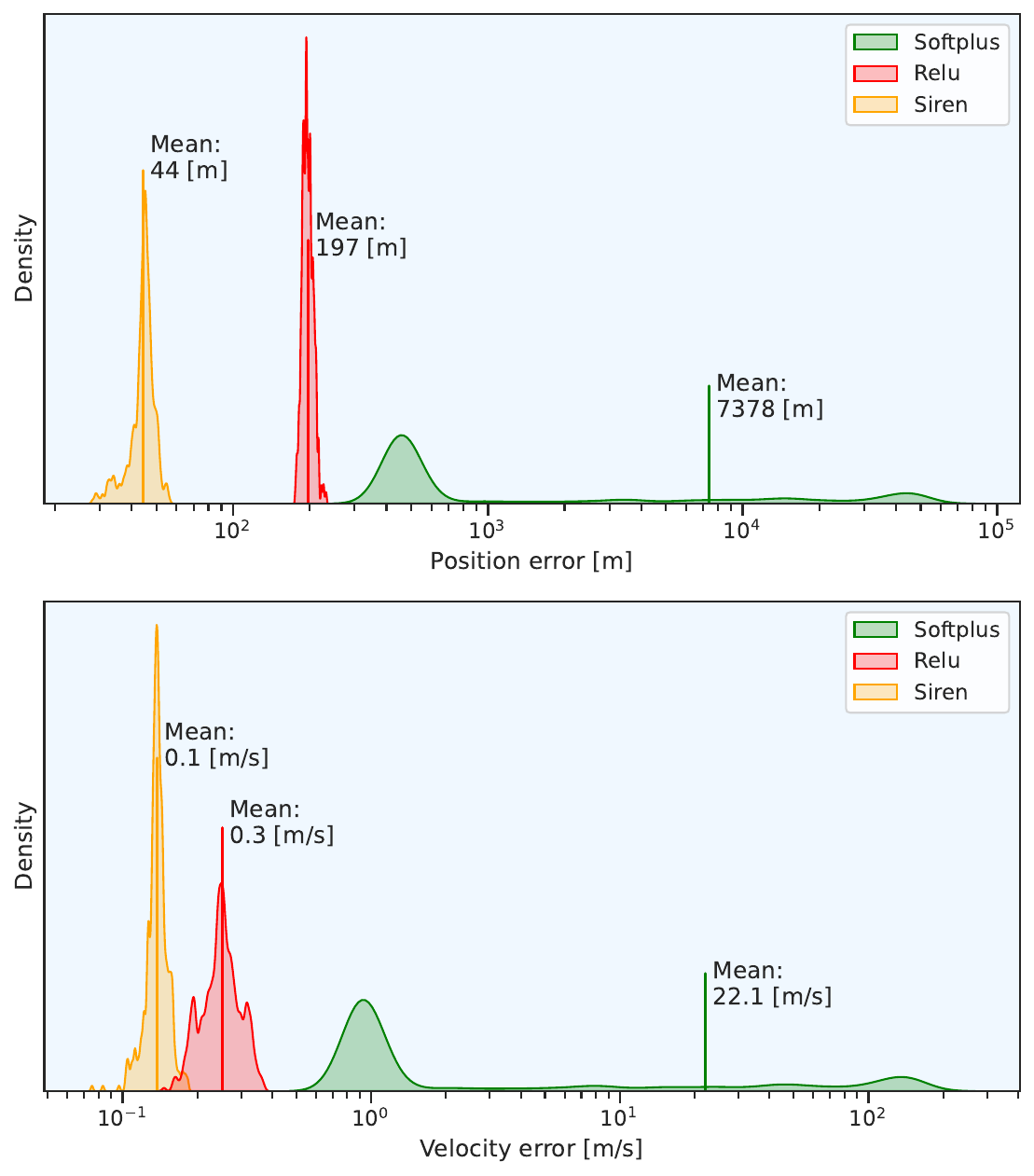}
  \caption{Asteroid landing: Final position and velocity errors over 500 initial conditions from validation dataset.}
	\label{fig:landing_errors}
\end{figure}
Here again the Siren network reaches the lowest training loss. Instead of showing the loss figures again we follow a similar analysis as in \cite{origer2024closing}, where we evaluate the trained G\&CNETs by propagating the system from a set of 500 initial conditions until the corresponding optimal time-of-flight has elapsed. We report the distribution of the final position and velocity errors to the target landing site in Fig.\ref{fig:landing_errors}. Here the ReLU network is clearly better suited than the Softplus as it allows a more accurate representation of the discontinuous throttle input. However the SIREN network overall achieves the best final mean position and velocity errors. The full version of this paper will include a comparison with the technique called 'Neural ODE fix', presented in \cite{origer2024closing}, which is an alternative way to improve the accuracy of G\&CNETs.
\subsection{Interplanetary transfer}\label{subse:transfer}
Let the radius of the target circular orbit be $R$ \cite{dario_seb_gcnet,origer2024closing}.
We define $\mathcal F = [\hat{\mathbf i}, \hat{\mathbf j}, \hat{\mathbf k}]$ to be a rotating frame of angular velocity $\boldsymbol \Omega= \sqrt{\mu/R^3} \hat {\mathbf k}$.
Thus, the target body's position $R\hat{\mathbf i}$ remains stationary in $\mathcal F$.
The equations of motion are as follows:
\begin{equation}
\label{eq:dyn_transfer}
\left\{ 
\begin{array}{l}
    \dot{x} = v_x  \\
    \dot{y} = v_y \\
    \dot{z} = v_z \\
    \dot{v}_x = -\frac{\mu}{r^3}x + 2 \Omega v_y + \Omega^2 x +\Gamma i_x \\
    \dot{v}_y = -\frac{\mu}{r^3}y - 2 \Omega v_x + \Omega^2 y +\Gamma i_y \\
    \dot{v}_z = -\frac{\mu}{r^3}z +\Gamma i_z
\end{array}
\right.
\end{equation}
The state vector $\mathbf{x}$ includes the position $\mathbf{r}=[x,y,z]$ and velocity $\mathbf{v}=[v_x,v_y,v_z]$, both defined in the rotating frame $\mathcal F$.
Here $r = \sqrt{x^2+y^2+z^2}$ and $\mu$ is the gravitational constant of the Sun.
The system is controlled by the thrust direction, described by the unit vector $\mathbf{\hat{t}} = [i_x, i_y, i_z]$, which generates an acceleration of magnitude $\Gamma$.
This time-optimal control problem aims to find a (piece-wise continuous) function for $\mathbf{\hat{t}}(t)$ and the optimal time-of-flight $t_f$, where $t\in [t_0,t_f]$, such that, under the dynamics described by Eq.\ref{eq:dyn_transfer}, the state is steered from any initial state $\mathbf{r}_0$, $\mathbf{v}_0$ to the desired target state $\mathbf{r}_t=R\mathbf{\hat{i}}$, $\mathbf{v}_t=\mathbf{0}$. 
We are thus minimizing the following cost function:
$J = t_f-t_0$ \cite{dario_seb_gcnet}. We also solve this optimal control problem with an indirect method \cite{pontryagin} (the details \cite{origer2024closing} are left out here for brevity). A portion of the training data is shown in App.\ref{app:3}.

For all G\&CNETs we use 400,000 optimal trajectories sampled at 100 points. We train the networks for 300 epochs with a training batchsize of 4096 optimal state-action pairs $(\mb{x}^*, \mb{u}^*)$.
The G\&CNETs are composed of an input layer which takes in 6 inputs (the state vector of the spacecraft), three hidden layers with 128 neurons each, and a linear output layer. The resulting G\&CNET has 34,307 parameters in total.
Finally, we use as the loss function $\mathcal L= 1-\frac{\mathbf{\hat{t}}^*\cdot \mathbf{\hat{t}}_{nn}}{\mathbf{\vert\hat{t}}^*\vert\vert \mathbf{\hat{t}}_{nn}\vert}$, minimizing the cosine similarity of the estimated thrust direction $\mathbf{\hat{t}}_{nn}$ and the ground truth $\mathbf{\hat{t}}^*$.

\begin{figure}[h]
  \centering
  \includegraphics[width=\columnwidth]{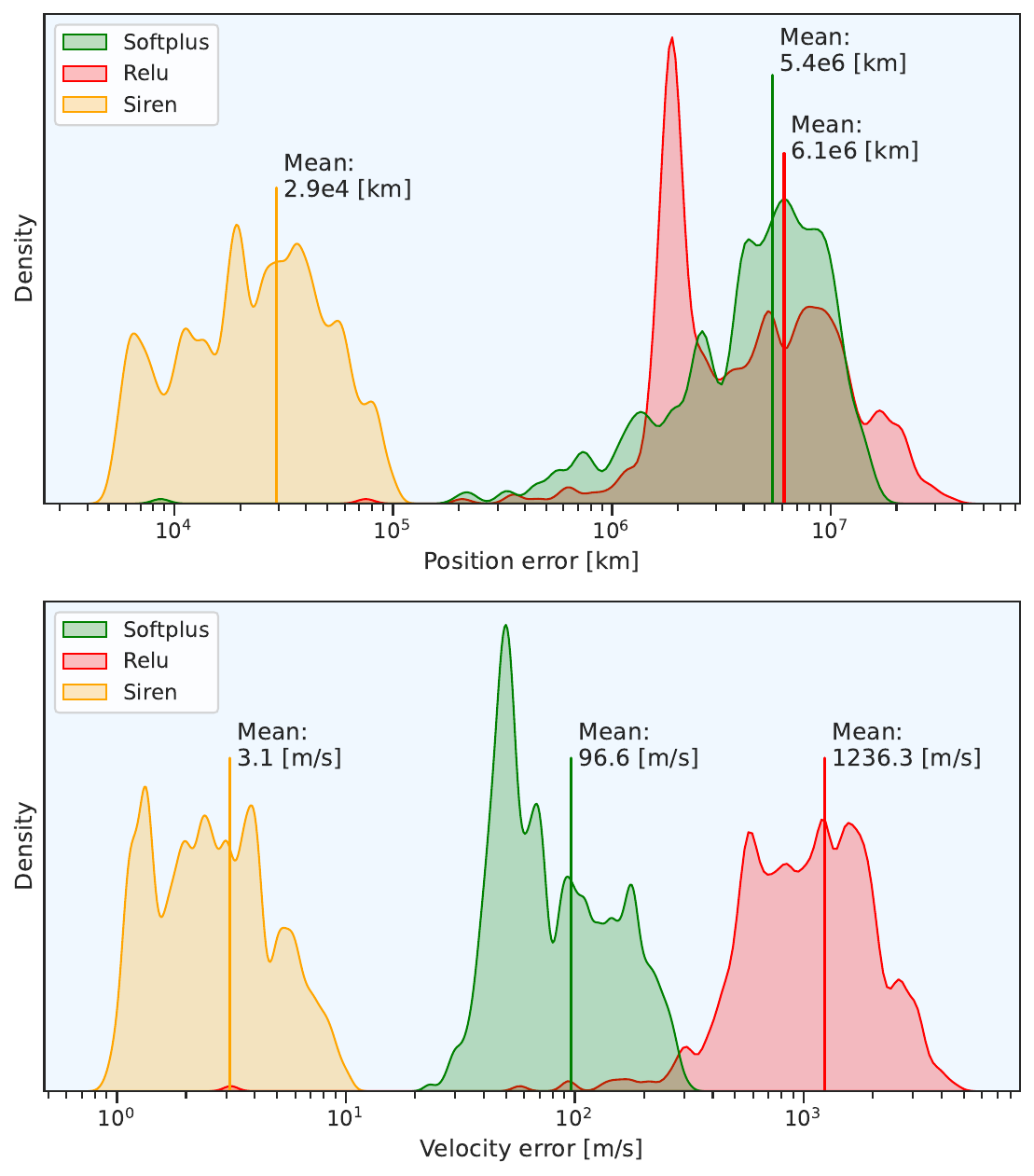}
  \caption{Interplanetary transfer: Final position and velocity errors over 500 initial conditions from validation dataset.}
	\label{fig:interplanetary_errors}
\end{figure}
Here again the Siren network reaches the lowest training loss. As before we evaluate the trained G\&CNETs by propagating the system from a set of 500 initial conditions until the corresponding optimal time-of-flight has elapsed. 
The distribution of the final position and velocity errors to the target rendezvous is shown in Fig.\ref{fig:interplanetary_errors}. Here the Softplus network performs better than the ReLU network, resulting in one order of magnitude lower final velocity errors. However the SIREN network overall achieves the best final mean position and velocity errors. 
We also found that for this optimal control problem a SIREN network with only 8,963 parameters is on par with a Softplus network with 1,479,103 parameters in terms of final position and velocity errors.

\section{Discussion}
The use of the sinusoidal representation networks (SIRENs) architecture for Guidance \& Control Networks (G\&CNETs) has been studied. We consider three optimal control tasks, each complex in their own way: drone racing, asteroid landing and an interplanetary transfer. In all cases, we find that switching to the SIREN architecture results in a significantly lower training loss compared to the typical G\&CNET architectures found in the literature. In addition, for the two latter problems, we find that SIRENs result in much lower final position and velocity errors the to target states.

While a full analysis of the reasons behind the superior performance of G\&CNETs with periodic activation functions is outside the scope of this short paper, it is planned for inclusion in the extended paper for the proceedings. One intuitive explanation for the observed results is that the sine activation function enables the network to harness inherent oscillations in the input space, effectively decomposing it into its Fourier spectrum. 

\printbibliography
\addcontentsline{toc}{section}{References}

\section*{Appendix}
\appendix
\section{Drone racing model}\label{app:1}
\begin{figure}[h!]
  \centering
  \includegraphics[width=0.6\columnwidth]{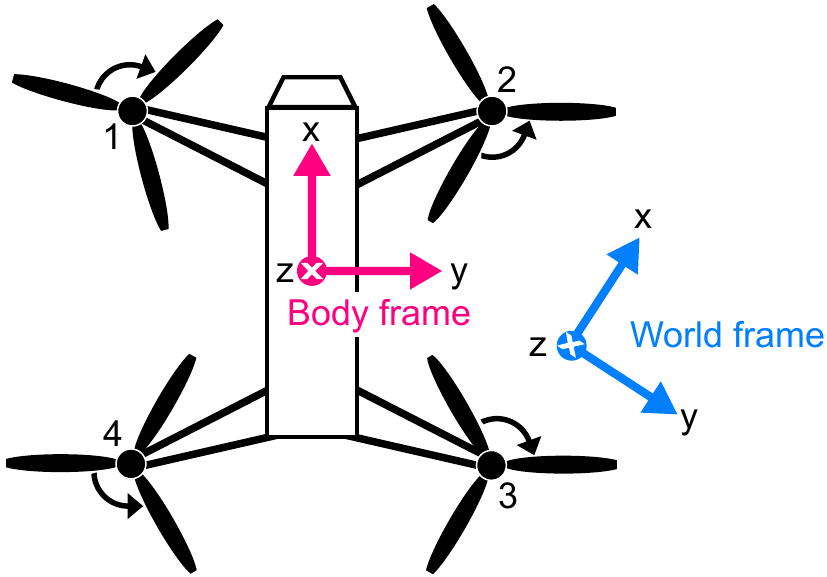}
  \caption{Coordinate frames (Body x-axis points to the front of the drone) \cite{origer2023guidance}.\vspace{-0mm}}
  \label{fig:coordframes}
\end{figure}

The coordinate frames used for the drone are shown in Fig.\ref{fig:coordframes}

Note that the rotational matrix $R(\mb{\lambda})$ transforms from the body to the world frame. We use the notation $c_{\theta}$ and $s_{\phi}$ to denote the cosine and sine of the corresponding Euler angle, respectively:
\begin{equation*}\label{eq:Euler_To World}
R(\mb{\lambda}) = 
\begingroup 
\setlength\arraycolsep{2pt}
\begin{bmatrix}
        c_{\theta}c_{\psi} & -c_{\phi}s_{\psi}+s_{\phi}s_{\theta}c_{\psi} & s_{\phi}s_{\psi}+c_{\phi}s_{\theta}c_{\psi} \\
        c_{\theta}s_{\psi} & c_{\phi}c_{\psi}+s_{\phi}s_{\theta}s_{\psi} & -s_{\phi}c_{\psi}+c_{\phi}s_{\theta}s_{\psi} \\
        -s_{\theta} & s_{\phi}c_{\theta} & c_{\phi}c_{\theta}
\end{bmatrix}
\endgroup
\end{equation*}
and $Q(\mb{\lambda})$ is the inverse transformation matrix: 
\begin{equation*} \label{eq:WorldToEuler}
Q(\mb{\lambda}) = 
\begin{bmatrix}
    1 & \sin{\phi} \tan{\theta}     & \cos{\phi} \tan{\theta} \\
    0 & \cos{\phi}                  & -\sin{\phi} \\
    0 & \sin{\phi} / \cos{\theta}   & \cos{\phi} / \cos{\theta}
\end{bmatrix}
\end{equation*}
The forces $\mb{F} = [F_x, F_y, F_z]^T$ are computed using the thrust and drag model from \cite{thrust_and_drag_model}. Note that the superscript $\square^B$ denotes the body frame, all model parameters are taken from \cite{origer2023guidance}.
\begin{align} \label{eq:F_model}
\begin{split}
    F_x &= - k_x v^B_x \sum_{i=1}^4 \omega_i \quad
    F_y = - k_y v^B_y \sum_{i=1}^4 \omega_i\\
    F_z &= -k_\omega \sum_{i=1}^4 \omega_i^2 - k_z v^B_z \sum_{i=1}^4 \omega_i - k_h (v^{B2}_x + v^{B2}_y)
\end{split}
\end{align}
and the moments $\mb{M}=[M_x, M_y, M_z]^T$ are defined as:

\begin{equation} \label{eq:M_model}
\begin{split}
    M_x &= k_p (\omega_1^2 - \omega_2^2 - \omega_3^2 + \omega_4^2) + k_{pv} v^{B}_y\\
    M_y &= k_q (\omega_1^2 + \omega_2^2 - \omega_3^2 - \omega_4^2) + k_{qv} v^{B}_x\\
    M_z &= k_{r1} (-\omega_1 + \omega_2 - \omega_3 + \omega_4) \\
    & + k_{r2} (-\dot{\omega}_1 + \dot{\omega}_2 - \dot{\omega}_3 + \dot{\omega}_4)  - k_{rr} r\\
\end{split}
\end{equation}
\section{Asteroid landing}\label{app:2}
\begin{figure}[h!]
   \centering
\includegraphics[width=\columnwidth]{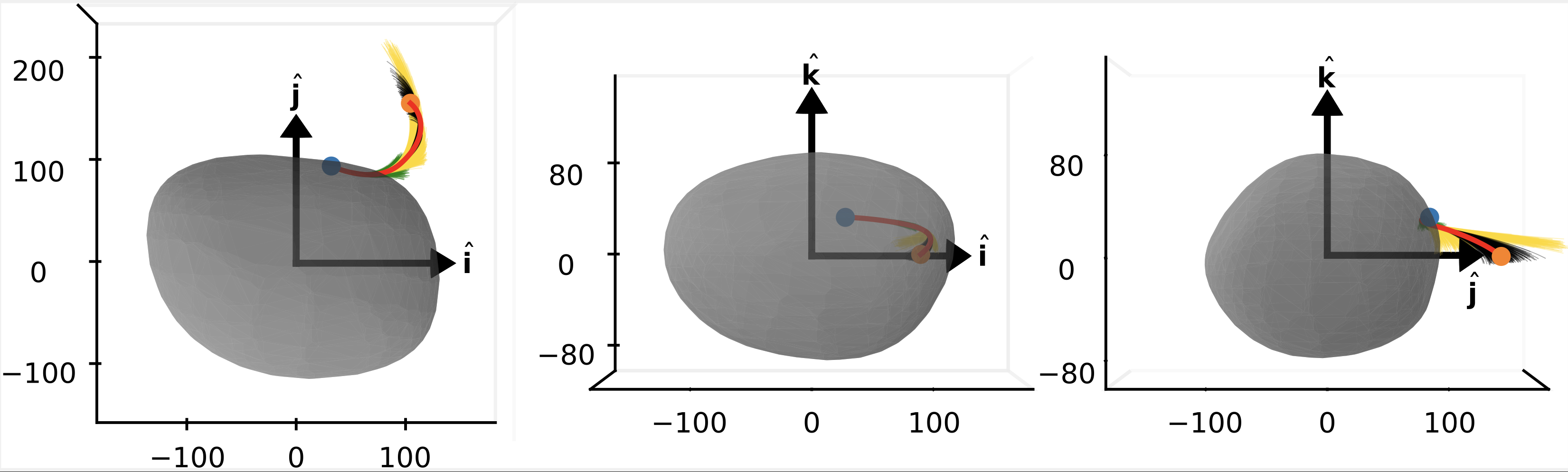}
\caption{Bundle of 2,000 optimal trajectories from training dataset \cite{origer2024closing}. Landing on Psyche shown in rotating frame $\mathcal R$. Axis unit is km.}
\label{fig:bundle_landing}
\end{figure}

A portion of the training dataset is shown in Fig.\ref{fig:bundle_landing}.
\section{Interplanetary transfer}\label{app:3}
\begin{figure}[h!]
   \centering
\includegraphics[width=\columnwidth]{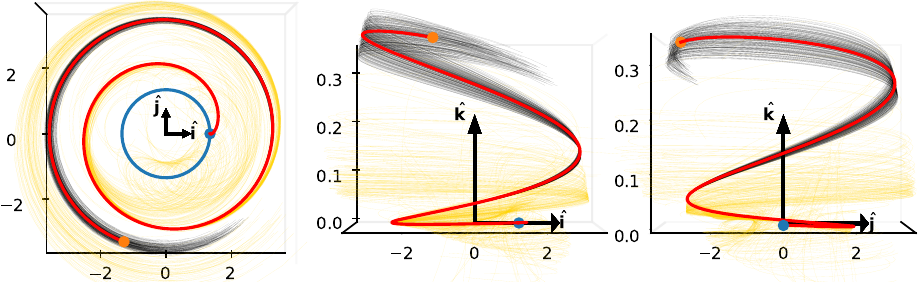}
\caption{Bundle of 400 optimal trajectories from training dataset \cite{origer2024closing}. Interplanetary transfer shown in rotating frame $\mathcal F$. Axis unit is AU.}\label{fig:bundle_transfer}
\end{figure}

A portion of the training dataset is shown in Fig.\ref{fig:bundle_transfer}.

\end{document}